%% file: main.tex
\documentclass[10pt,twocolumn,letterpaper]{article}

\usepackage{cvpr}              

\usepackage{graphicx}
\usepackage{amsmath}
\usepackage{amssymb}
\usepackage{booktabs}
\usepackage{colortbl}
\usepackage{enumitem}
\usepackage{arydshln} 
\usepackage[pagebackref,breaklinks,colorlinks]{hyperref}

\newcommand{\modelname}{MobileR2L}

\usepackage[capitalize]{cleveref}
\crefname{section}{Sec.}{Secs.}
\Crefname{section}{Section}{Sections}
\Crefname{table}{Table}{Tables}
\crefname{table}{Tab.}{Tabs.}

\newcommand*{\affmark}[1][*]{\textsuperscript{#1}}
\definecolor{urlcolor}{rgb}{0.93,0.01,0.55}

\begin{document}

\title{Real-Time Neural Light Field on Mobile Devices}


\author{Junli Cao\affmark[1]\quad\quad
\quad  Huan Wang\affmark[2]\quad\quad Pavlo Chemerys\affmark[1] \quad\quad
Vladislav Shakhrai\affmark[1]\quad\quad  Ju Hu\affmark[1]\\ Yun Fu\affmark[2] \quad\quad Denys Makoviichuk\affmark[1] \quad\quad Sergey Tulyakov\affmark[1] \quad\quad Jian Ren\affmark[1]\\
{\affmark[1]Snap Inc.\quad\quad\quad\affmark[2]Northeastern University }
}
\maketitle

\input{section/0_abstract}
\input{section/1_intro_2}

\input{section/2_related}
\input{section/3_method}

\input{section/4_exp}
\input{section/5_conclusion}


{\small
\bibliographystyle{ieee_fullname}
\bibliography{egbib}
}
\clearpage
\appendix
\input{section/6_supp}
\end{document}

%% file: section/0_abstract.tex
\begin{abstract}
Recent efforts in Neural Rendering Fields (NeRF) have shown impressive results on novel view synthesis by utilizing implicit neural representation to represent 3D scenes.
Due to the process of volumetric rendering, the inference speed for NeRF is extremely slow, limiting the application scenarios of utilizing NeRF on resource-constrained hardware, such as mobile devices. 
Many works have been conducted to reduce the latency of running NeRF models. 
However, most of them still require high-end GPU for acceleration or extra storage memory, which is all unavailable on mobile devices. 
Another emerging direction utilizes the neural light field (NeLF) for speedup, as only one forward pass is performed on a ray to predict the pixel color. 
Nevertheless, to reach a similar rendering quality as NeRF, the network in NeLF is designed with intensive computation, which is not mobile-friendly.
In this work, we propose an efficient network that runs in real-time on mobile devices for neural rendering. 
We follow the setting of NeLF to train our network. 
Unlike existing works, we introduce a novel network architecture that runs efficiently on mobile devices with low latency and small size, \emph{i.e.}, saving $15\times \sim 24\times$ storage compared with MobileNeRF. 
Our model achieves high-resolution generation while maintaining real-time inference for both synthetic and real-world scenes on mobile devices, \emph{e.g.,} $18.04$ms (iPhone 13) for rendering one $1008\times756$ image of real 3D scenes.  
Additionally, we achieve similar image quality as NeRF and better quality than MobileNeRF (PSNR $26.15$ \emph{vs.} $25.91$ on the real-world forward-facing dataset)\footnote{More demo examples in our~\href{https://snap-research.github.io/MobileR2L/}{\color{urlcolor}{Webpage}}.}. 
\end{abstract}

%% file: section/1_intro_2.tex
\section{Introduction}
\input{figs/0_teaser}

Remarkable progress seen in the domain of neural rendering~\cite{nerf} promises to democratize asset creation and rendering, where no mesh, texture, or material is required -- only a neural network that learns a representation of an object or a scene from multi-view observations. The trained model can be queried at arbitrary viewpoints to generate novel views. To be made widely available, this exciting application requires such methods to run on resource-constrained devices, such as mobile phones, conforming to their limitations in computing, wireless connectivity, and hard drive capacity. 

Unfortunately, the impressive image quality and capabilities of NeRF~\cite{nerf} come with a price of slow rendering speed. To return the color of the queried pixel, hundreds of points need to be sampled along the ray that ends up in that pixel, which is then integrated to get the radiance. To enable real-time applications, many works have been proposed~\cite{kilonerf,plenoctrees,plenoxels,instant-ngp}, yet, they still require high-end GPUs for rendering and hence are not available for resource-constrained applications on mobile or edge devices. An attempt is made to trade rendering speed with storage in MobileNeRF~\cite{mobilenerf}. While showing promising acceleration results, their method requires storage for texturing saving. For example, for a single real-world scene from the forward-facing dataset~\cite{nerf}, MobileNeRF requires $201.5$MB of storage. 
Clearly, downloading and storing tens, hundreds, or even thousands of such scenes in MobileNeRF format on a device is prohibitively expensive. 

A different approach is taken in Neural Light Fields (NeLF) that directly maps a ray to the RGB color of the pixel by performing only one forward pass per ray, resulting in faster rendering speed~\cite{levoy1996light,attal2022learning,liu2022neulf,r2l}. Training NeLF is challenging and hence requires increased network capacity. For example, Wang \textit{et al.}~\cite{r2l} propose an $88$-layer fully-connected network with residual connections to distill a pre-trained radiance model effectively. While their approach achieves better rendering results than vanilla NeRF at $30\times$ speedup, running it on mobile devices is still not possible, as it takes three seconds to render one $200\times200$ image on iPhone 13 shown in our experiments.

In this work, we propose \modelname, a real-time neural rendering model built with mobile devices in mind. Our training follows a similar distillation procedure introduced in R2L~\cite{r2l}. Differently, instead of using an MLP, a backbone network used by most neural representations, we show that a well-designed \emph{convolutional} network can achieve real-time speed with the rendering quality similar to MLP. In particular, we revisit the network design choices made in R2L and propose to use the $1\times1$ Conv layer in the backbone. A further challenge with running a NeRF or NeLF on mobile devices is an excessive requirement of RAM. For example, to render an $800\times800$ image, one needs to sample $640,000$ rays that need to be stored, causing out-of-memory issues. In 3D-aware generative models~\cite{eg3d,stylenerf,headnerf}, this issue is alleviated by rendering a radiance feature volume and upsampling it with a convolutional network to obtain a higher resolution. Inspired by this, we render a light-field volume that is upsampled to the required resolution.  
Our \modelname~features several major advantages over existing works:
\begin{itemize}[leftmargin=1em]
\setlength\itemsep{-0.5em}
    \item \modelname~achieves real-time inference speed on mobile devices (Tab.~\ref{table:speed}) with better rendering quality, \emph{e.g.}, PSNR, than MobileNeRF on the synthetic and real-world datasets (Tab.~\ref{table:quantitative}).
    \item \modelname~requires an order of magnitude less storage, reducing the model size to $8.3$MB, which is ${15.2\times\sim24.3\times}$ less than MobileNeRF.
\end{itemize}
Due to these contributions, \modelname~can unlock wide adoption of neural rendering in real-world applications on mobile devices, such as a virtual try-on, where the real-time interaction between devices and users is achieved (Fig.~\ref{fig:teaser}).

%% file: figs/0_teaser.tex
\begin{figure}[ht]
  \includegraphics[width=1.0\linewidth]{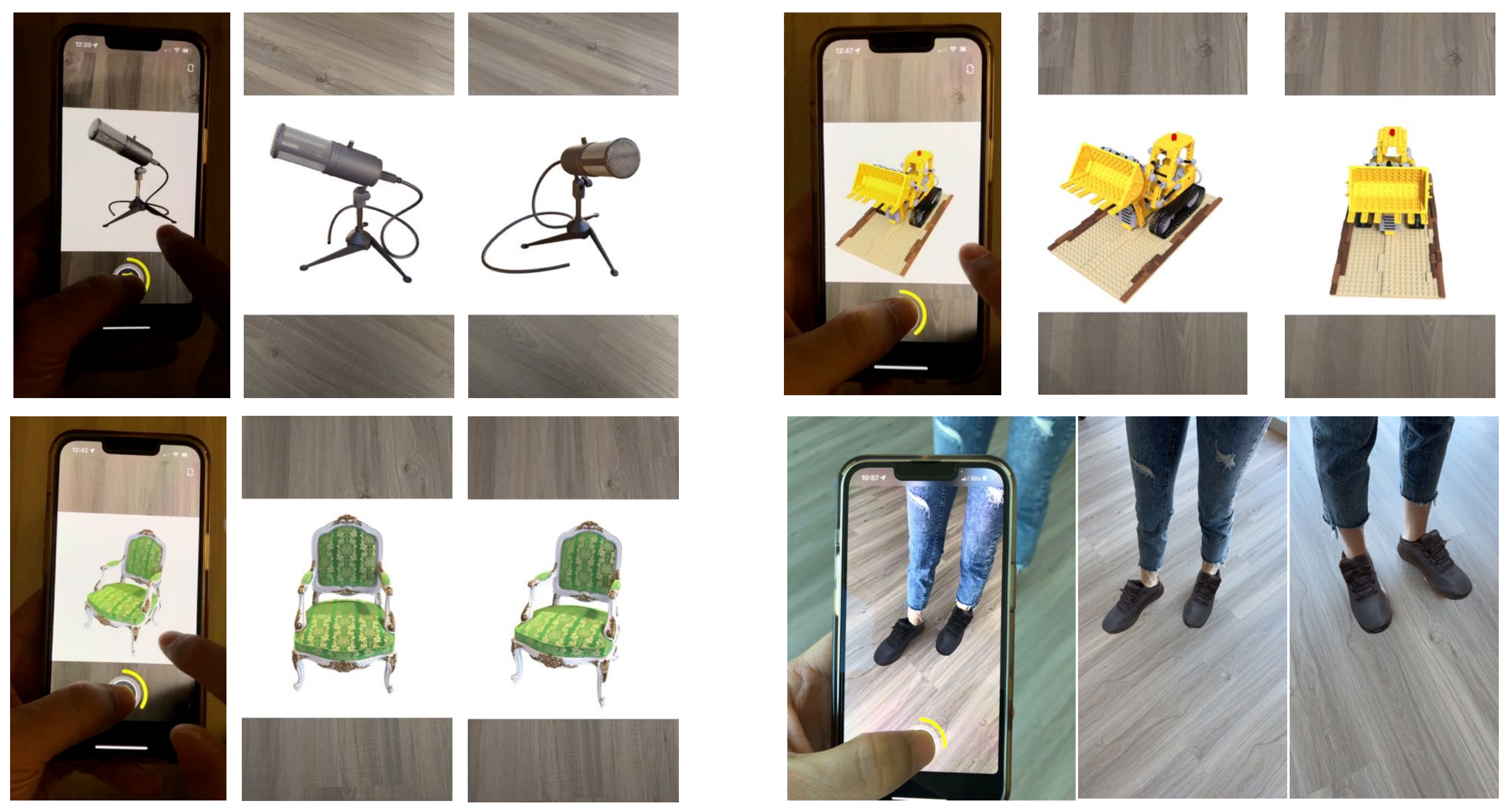}  
  \caption{Examples of deploying our approach on mobile devices for real-time interaction with users. Due to the small model size ($8.3$MB) and fast inference speed ($18\sim26$ms per image on iPhone 13), we can build neural rendering applications where users interact with 3D objects on their devices, enabling various applications such as virtual try-on. We use publicly available software to make the on-device application for visualization~\cite{lens-studio, Snap-ML}.}
  \label{fig:teaser}
\end{figure}

%% file: section/2_related.tex
\section{Related Work}
\noindent \textbf{Neural Radiance Field (NeRF).} NeRF~\cite{nerf} shows the possibility of representing a scene with a simple multi-layer perceptron (MLP) network. Going forward, many extensions follow up in improving rendering quality (\eg, MipNeRF~\cite{mip-nerf}, MipNeRF 360~\cite{mip-nerf360}, and Ref-NeRF~\cite{verbin2022ref}), rendering efficiency (\eg, NSVF~\cite{liu2020neural}, Nex~\cite{wizadwongsa2021nex}, AutoInt~\cite{lindell2021autoint}, FastNeRF~\cite{garbin2021fastnerf}, Baking~\cite{hedman2021baking}, Plenoctree~\cite{plenoctrees}, KiloNeRF~\cite{kilonerf}, DeRF~\cite{rebain2021derf}, DoNeRF~\cite{neff2021donerf}, R2L~\cite{r2l}, and MobileNeRF~\cite{mobilenerf}) and training efficiency (\eg, Plenoxels~\cite{plenoxels}, and Instant-NGP~\cite{instant-ngp}).

\noindent \textbf{Efficient NeRF Rendering.}
Since this paper falls into the category of improving \textit{rendering efficiency} as we target \textit{real-time} rendering on \textit{mobile} devices, we single out the papers of this group and discuss them in length here. There are generally four groups.
\textbf{(1)} The first group trades speed with space, \ie, they precompute and cache scene representations and the rendering reduces to table lookup. Efficient data structure like sparse octree, \eg, Plenoctree~\cite{plenoctrees}, is usually utilized to make the rendering even faster. 
\textbf{(2)} The second attempts reduce the number of sampled points along the camera ray during rendering as it is the root cause of prohibitively slow rendering speed. Fewer sampled points typically lead to performance degradation, so as compensation, they usually introduce extra information, such as depth, \eg, DoNeRF~\cite{neff2021donerf}, or mesh, \eg, MobileNeRF~\cite{mobilenerf}, to maintain the visual quality.
\textbf{(3)} The third group takes a ``divide and conquer'' strategy. DeRF~\cite{rebain2021derf} decomposes the scene spatially to Voronoi diagrams and learns each diagram with a small network. KiloNeRF~\cite{kilonerf} also employs a decomposition scheme. Differently, they decompose the scene into thousands of small regular grids. Each is addressed with a small network. Such decomposition poses challenges to parallelism. Thus they utilize customized parallelism implementation to obtain speedup.
\textbf{(4)} The fourth group is a newly surging one, represented by the recent works RSEN~\cite{attal2022learning} and R2L~\cite{r2l}. They achieve rendering efficiency by representing the scene with \textit{NeLF (neural light field}) instead of NeRF. NeLF avoids the dense sampling on camera ray, resulting in a much faster rendering speed than NeRF. On the downside, NeLF is typically much harder to learn than NeRF. As a remedy, these works~(such as \cite{r2l}) typically integrate a pre-trained NeRF model as a teacher to synthesize additional pseudo data for \textit{distillation}~\cite{bucilua2006model,hinton2015distilling}. Therefore, the resulting model is fast with a reasonably small representation size, \ie, the model size.   

\noindent \textbf{Neural Light Field (NeLF).} Light field is a different way of representing scenes. The idea has a long history in the computer graphics community, \eg, Light fields~\cite{levoy1996light} and Lumigraphs~\cite{gortler1996lumigraph} cache plenty of images and enable real-time rendering at the cost of limited camera pose and excessive storage overhead. One of the most intriguing properties of NeLF is that rendering one image only requires one network forward, resulting in a significantly faster speed than NeRF-based methods. With the recent surge of the neural radiance field, some works attempt to revive the idea of the neural light field for efficient neural rendering. Sitzmann \emph{et al.}~\cite{sitzmann2021light} materialize the idea of using a neural network to model the scene, and the rendering process reduces to a single network forward. Despite the encouraging idea, their method has only been evaluated on scenes with simple shapes, not matching the quality of NeRF on complex real-world scenes. Later, RSEN~\cite{attal2022learning} and R2L~\cite{r2l} are introduced. RSEN divides the space into many voxel grids. Only in each grid, it is a NeLF, which needs alpha-composition to render the final color, making their method \textit{a mixture of NeLF and NeRF}. R2L~\cite{r2l} is a pure NeLF network that avoids the alpha-composition step in rendering, which is also one of the most relevant works to this paper. However, R2L is still not compact and fast enough for mobile devices. Based on our empirical study, an R2L model runs for around three seconds per frame on iPhone 13 even for low-resolution like $200\times200$. This paper is meant to push the NeRF-to-NeLF idea even further, making it able to perform \textit{real-time} rendering on mobile devices.

We will mainly compare to MobileNeRF~\cite{mobilenerf} in this paper as it is the \textit{only} method, to our best knowledge, that can run on mobile devices with matching quality to NeRF~\cite{nerf}.

%% file: section/3_method.tex
\section{Methodology}
\subsection{Prerequisites: NeRF and R2L}
\noindent\textbf{NeRF.} NeRF~\cite{nerf} represents the scene implicitly with an MLP network $F_{\Theta}$, which maps the $5$D coordinates (spatial location $(x, y, z)$ and view direction $(\theta, \phi)$) to a $1$D volume density (opacity, denoted as $\sigma$ here) and $3$D radiance (denoted as $\mathbf{c}$) such that $F_{\Theta}: \mathbb{R}^5  \mapsto \mathbb{R}^4$. Each pixel of an image is associated with a camera ray. To predict the color $\hat{C}$ of a pixel, the NeRF method samples many points (denoted as $N$ below) along the camera ray and accumulates the radiance $\mathbf{c}$ of all these points via \textit{alpha compositing}~\cite{kajiya1984ray,max1995optical,nerf}:
\begin{equation}
\small
\begin{aligned}
    \hat{C}(\mathbf{r}) &= \sum_{i=1}^N T_i \cdot (1 - \exp(-\sigma_i \delta_i)) \cdot \mathbf{c}_i, \\
    (\mathbf{c}_i, \sigma_i) &= F_{\Theta}(\mathbf{r}(t_i), \mathbf{d}), \\
    T_i &= \exp(-\sum_{j=1}^{i-1} \sigma_j \delta_j),
\end{aligned}
\label{eq:alpha_compositing}
\end{equation}
where $\mathbf{r}$ means the camera ray; $\mathbf{r}(t_i) = \mathbf{o} + t_i\mathbf{d}$ represents the location of a point on the ray with origin $\mathbf{o}$ and direction $\mathbf{d}$; $t_i$ is the Euclidean distance, \ie, a scalar, of the point away from the origin; and $\delta_i = t_{i+1} - t_i$ refers to the distance between two adjacent sampled points. A stratified sampling approach is employed in NeRF~\cite{nerf} to sample the $t_i$ in Eqn.~\ref{eq:alpha_compositing}. To enrich the input information, the position and direction coordinates are encoded by \textit{positional encoding}~\cite{vaswani2017attention}, which maps a scalar ($\mathbb{R}$) to a higher dimensional space ($\mathbb{R}^{2L+1}$) through cosine and sine functions, where $L$ (a predefined constant) stands for the frequency order (in the original NeRF~\cite{nerf}, $L=10$ for positional coordinates and $L=4$ for direction coordinates).

The whole formulation and training of NeRF are straightforward. One critical problem preventing fast inference in NeRF is that the $N$, \ie, the number of sampled points, in Eqn.~\ref{eq:alpha_compositing} is pretty large ($256$ in the original NeRF paper due to their two-stage coarse-to-fine design). Therefore, the rendering computation for even a single pixel is prohibitively heavy. The solution proposed by R2L~\cite{r2l} is distilling the NeRF representation to NeLF.
\input{figs/6_train_pipeline}
\input{figs/1_architecture}

\noindent\textbf{R2L.} Essentially, a NeLF function maps the oriented ray to RGB.
To enrich the input information, R2L proposes a new ray representation -- they also sample points along the ray just like NeRF~\cite{nerf} does; but differently, they \textit{concatenate} the points to one vector, which is used as the ray representation and fed into a neural network to learn the RGB. Similar to NeRF, positional encoding~\cite{vaswani2017attention} is also adopted in R2L to map each scalar coordinate to a high dimensional space. During training, the points are \textit{randomly} (by a uniform distribution) sampled; during testing, the points are fixed.

The output of the R2L model is directly RGB, no density learned, and there is no extra alpha-compositing step, which makes R2L much faster than NeRF in rendering. One downside of the NeLF framework is, as shown in R2L~\cite{r2l}, the NeLF representation is much harder to learn than NeRF; so as a remedy, R2L proposes an $88$-layer deep ResMLP (residual MLP) architecture (much deeper than the network of NeRF) to serve as the mapping function.

R2L has two stages in training. In the first stage, they use a pre-trained NeRF model as a teacher to synthesize excessive \texttt{(origin, direction, RGB)} triplets as pseudo data; and then fed the pseudo data to train the deep ResMLP. This stage can make the R2L model achieve comparable performance to the teacher NeRF model. In the second stage, they finetune the R2L network from the first stage on the \textit{original} data -- this step can further boost the rendering quality as shown in the R2L work~\cite{r2l}.

\subsection{MobileR2L}
\subsubsection{Overview}
We follow the learning process of R2L to train our proposed \modelname, namely, using a pre-trained teacher model, such as NeRF~\cite{nerf}, to generate pseudo data for the training of a lightweight neural network. To reduce the inference speed, we aim only to forward the network \emph{once} when rendering an image. However, under the design of R2L, although one pixel only requires one network forward, directly feeding rays with large spatial size, \emph{e.g.}, $800\times800$, into a network causes memory issues. Therefore, R2L forwards a partial of rays each time, increasing the speed overhead.
To solve the problem, we introduce super-resolution modules, which upsample the low-resolution input, \emph{e.g.}, $100\times100$, to a high-resolution image. 
Thus, we can obtain a high-resolution image with only one forward pass of the neural network during inference time. The training and inference pipeline is illustrated in Fig.~\ref{fig:train}, and we introduce more details for our network architecture in the following. 

\subsubsection{Network Architectures}
The input rays can be represented as
$\mathbf{x}\in\mathbb{R}^{B,6,H,W}$, where $B$ denotes the batch size and $H$ and $W$ denote the spatial size. The ray origin and view directions are concatenated as the second dimension of $\mathbf{x}$. We then apply positional encoding $\gamma(\cdot)$ on $\mathbf{x}$ to map the ray origin and view directions into a higher dimension. Thus, we get the input of our neural network as $\gamma(\mathbf{x})$.

The network includes two main parts: an efficient backbone  and Super-Resolution (SR) modules for high-resolution rendering, with the architecture provided in Fig.~\ref{fig:architecture}. Instead of using Fully Connected (FC) or linear layers for the network that is adopted by existing works~\cite{nerf, r2l}, we only apply convolution (CONV) layers in the backbone and super-resolution modules.

There are two main reasons for replacing FC with CONV layers. First, the CONV layer is better optimized by compilers than the FC layer~\cite{liu2018efficient}. Under the same number of parameters, the model with CONV $1\times1$ runs around $27\%$ faster than the model with FC layers, as shown in Tab.~\ref{table:ablation_1}. Second, suppose FC is used in the backbone, in that case, extra \texttt{Reshape} and \texttt{Permute} operations are required to modify the dimension of the output features from the FC to make the features compatible with the CONV layer in the super-resolution modules, as the FC and CONV calculate different tensor dimensions. However, such \texttt{Reshape} or \texttt{Permute} operation might not be hardware-friendly on some mobile devices~\cite{li2022efficientformer,li2022rethinking}. With the CONV employed as the operator in the network, we then present more details for the backbone and SR modules.

\noindent\textbf{Efficient Backbone.} The architecture of the backbone follows the design of residual blocks from R2L~\cite{r2l}. Different from R2L, we adopt the CONV layer instead of the FC layer in each residual block. The CONV layer has the kernel size and stride as $1$. Additionally, we use the normalization and activation functions in each residual block, which can improve the network performance without introducing latency overhead (see experimental details in Tab.~\ref{table:ablation_1}). The normalization and activation are chosen as batch normalization~\cite{ioffe2015batch} and GeLU~\cite{hendrycks2016gaussian}. The backbone contains $60$ CONV layers in total.

\noindent\textbf{Super-Resolution Modules.} To reduce the latency when running the neural rendering on mobile devices, we aim to forward the neural network \emph{once} to get the synthetic image. However, the existing network design of the neural light field requires large memory for rendering a high-resolution image, which surpasses the memory constraint on mobile devices. For example, rendering an image of $800\times800$ requires the prediction of $640,000$ rays. Forwarding these rays at once using the network from R2L~\cite{r2l} causes the out of memory issue even on the Nvidia Tesla A$100$ GPU ($40$G memory).

To reduce the memory and latency cost for high-resolution generation, instead of forwarding the number of rays that equals to the number of pixels, we only forward a partial of rays and learn all the pixels via super-resolution. Specifically, we propose to use the super-resolution modules following the efficient backbone to upsample the output to a high-resolution image. For example, to generate a $800\times800$ image, we forward a $4$D tensor $\mathbf{x}$ with spatial size as $100\times100$ to the network and upsample the output from backbone three times (\textit{i.e.},~upsample by 2$\times$ each time). The SR module includes two stacked residual blocks. The first block includes three CONV layers with one as a $2$D Transpose CONV layer and two CONV $1\times1$ layers; the second block includes two CONV $1\times1$ layers. After the SR modules, we apply another CONV layer followed by the Sigmoid activation to predict the final RGB color. We denote our model as \textit{D$60$-SR$3$} where it contains $60$ CONV layers in the efficient backbone and $3$ SR modules.


%% file: figs/6_train_pipeline.tex
\begin{figure}[t!]
  \includegraphics[width=1\linewidth]{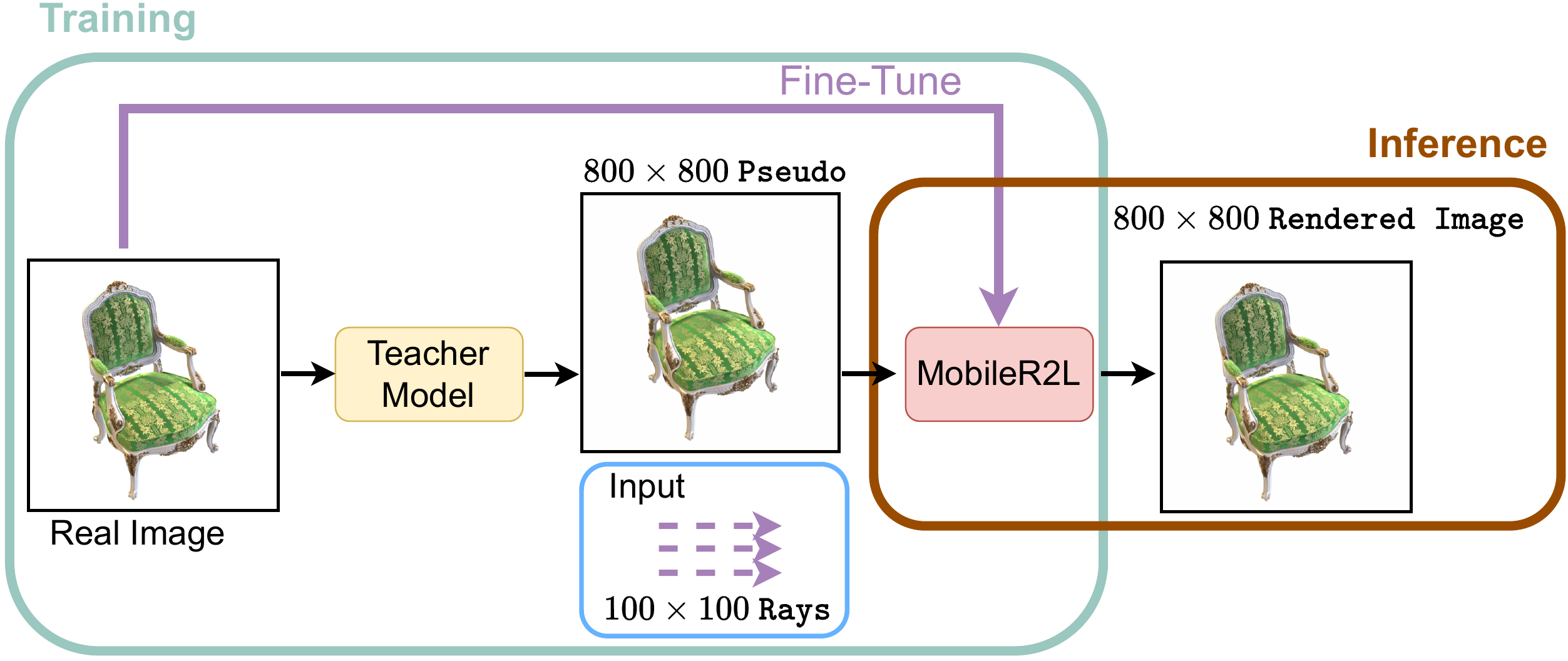} 
  \caption{\textbf{Training and Inference Pipeline.} 
  The training involves a teacher model to generate pseudo data, which is used to learn the \modelname.
  The teacher model, \emph{e.g.}, NeRF, is trained on real images. Once we have the teacher model, we use it to generate pseudo images, \emph{e.g.}, images with the resolution of $800\times800$, in addition to down-scaled rays, \emph{e.g.}, rays with spatial size as $100\times100$, that share the same origin with the pseudo images to train the \modelname. After that, we use the real data to fine-tune \modelname. For inference, we directly forward the rays into the pre-trained \modelname~to render images.}
  \label{fig:train}
  \end{figure}

%% file: figs/1_architecture.tex
\begin{figure*}[ht]
  \includegraphics[width=1\textwidth]{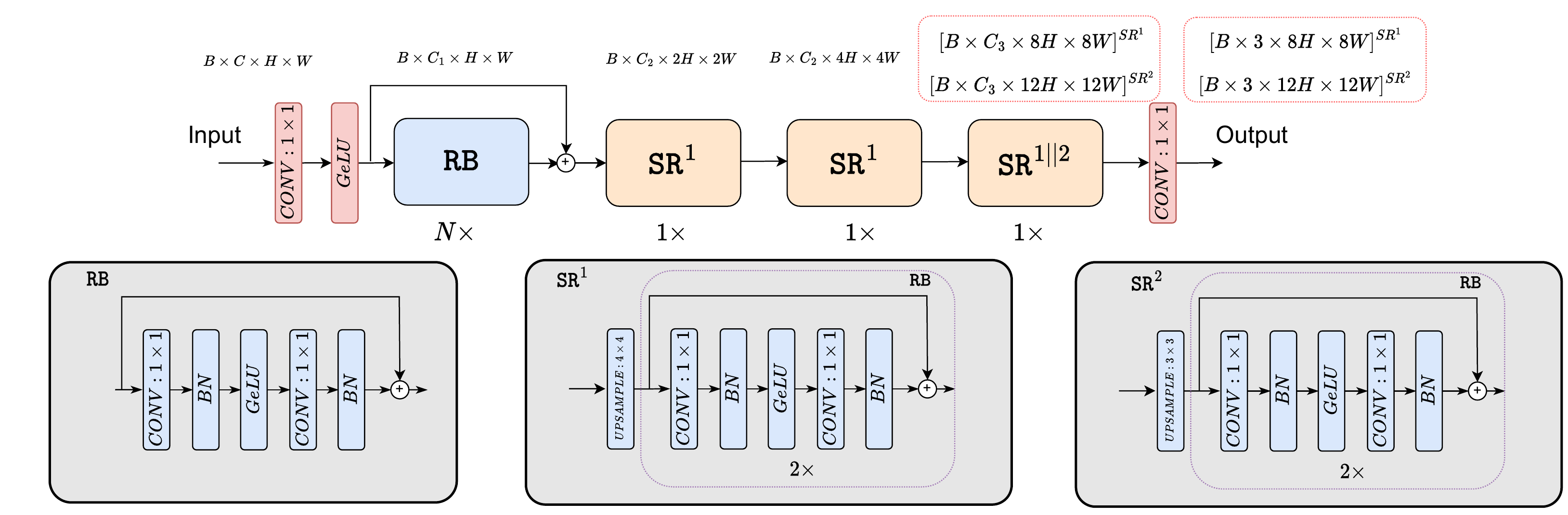}
  \caption{\textbf{Overview of Network.} The input tensor of \modelname~has 4D shape: batch, channel, height, and width. The backbone includes residual blocks (RB) that is repeated $28$ times ($N=28$). Following the backbone, there are two types of super-resolution (SR) modules. 
  The first SR module ({$\text{SR}^{1}$}) has kernerl size $4\times4$ in the Transpose CONV layer that doubles the input ${H, W}$ to ${2H, 2W}$, whereas the second SR module ({$\text{SR}^{2}$}) has kernerl size $3\times3$, tripling the spatial size to ${3H, 3W}$. The configuration of $3\times\text{SR}^{1}$ is used in the synthetic $360^{\circ}$  dataset that upsamples the input $8\times$. For the real-world forward-facing dataset, we use the combination of  $2\times\text{SR}^{1} + \text{SR}^{2}$ that upsamples the input $12\times$. Moreover, we use various output channels across {RB} and {SR}: $C_1 = 256$, $C_2 = 64$, and $C_3 = 16$.}
  \label{fig:architecture}
\end{figure*}

%% file: section/4_exp.tex
\section{Experiments}

\noindent\textbf{Datasets.} We conduct the comparisons mainly on two datasets: realistic synthetic $360^{\circ}$~\cite{nerf} and real-world forward-facing~\cite{nerf,mildenhall2019local}.  The synthetic $360^{\circ}$ dataset contains $8$ path-traced scenes, with each scene including $100$ images for training and $200$ images for testing. 
Forward-facing contains $8$ real-world scenes captured by cellphones, where the images in each scene vary from $20$ to $60$, and  $1/8$ images are used for testing.
We conduct our experiments (training and testing) on the resolution of $800\times800$ for synthetic $360^{\circ}$ and $1008\times756$ ($4\times$ down-scaled from the original resolution) for forward-facing.

\noindent\textbf{Implementation Details.} We follow the training scheme of R2L~\cite{r2l}, \emph{i.e.}, using a teacher model to render pseudo images for the training of \modelname~network. Specifically, we synthesize $10$K pseudo images from the pre-trained teacher model~\cite{NeRF-Factory} for each scene. 
We first train our \modelname~on the generated pseudo data and then fine-tune it on the real data, as shown in Fig. \ref{fig:train}. In all the experiments, we employ Adam~\cite{adam} optimizer with an initial learning rate $5\times10^{-4}$ that decays during the training. Our experiments are conducted on a cluster of Nvidia V100 and A100 GPUs with the batch size as $54$ for the main results on the synthetic $360^{\circ}$ and batch size as $36$ on the forward-facing dataset.

Different from R2L, the spatial size of the input rays and the output rendered images in our approach are different. For each high-resolution image generated by the teacher model, we save the input rays corresponding to a lower-resolution image where the camera origins and directions are the same as the high-resolution one while the focal length is down-scaled accordingly. Additionally, we do \textit{not} sample the rays from different images as in R2L. Instead, the rays in each training sample share the same origin and reserve their spatial locations. 

Considering the training data of the synthetic $360^{\circ}$ and forward-facing datasets have different resolutions, the spatial size of the inputs for the two datasets are slightly different.
Our network takes input with the spatial size of $100\times100$ for the synthetic $360^{\circ}$ dataset and upsamples by $8\times$ to render $800\times800$ RGB images. In contrast, the spatial size of $84\times63$ is used in the forward-facing dataset, and $1008 \times 756$ image ($12\times$ upsampling) is rendered. The kernel size and padding are adjusted in the last transposed  CONV layer to achieve $8\times$ and $12\times$ upsampling with the $3$ SR blocks.

\subsection{Comparisons}
\input{tabs/quantitative_comparison}
\begin{figure}[t]
\centering
\resizebox{1\linewidth}{!}{
\setlength{\tabcolsep}{0.5mm}
\begin{tabular}{ccccc}
\includegraphics[width=0.2\linewidth]{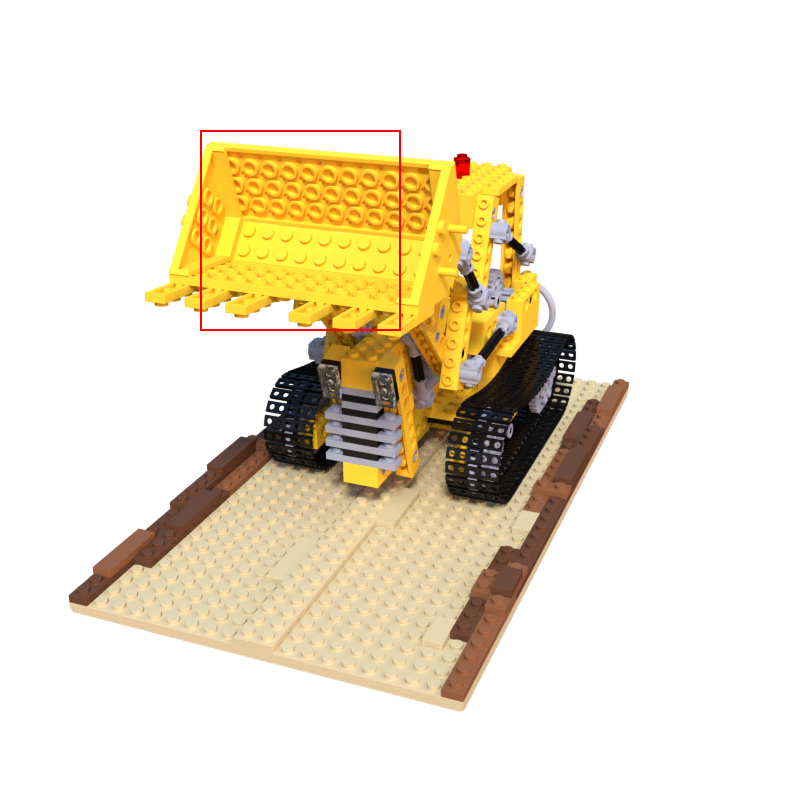} &
\includegraphics[width=0.2\linewidth]{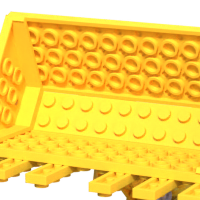} &
\includegraphics[width=0.2\linewidth]{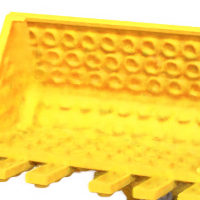} &
\includegraphics[width=0.2\linewidth]{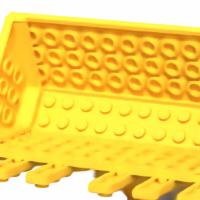} \\
\texttt{Lego} & (a) GT & (b) NeRF & (c) Ours \\
\includegraphics[width=0.267\linewidth]{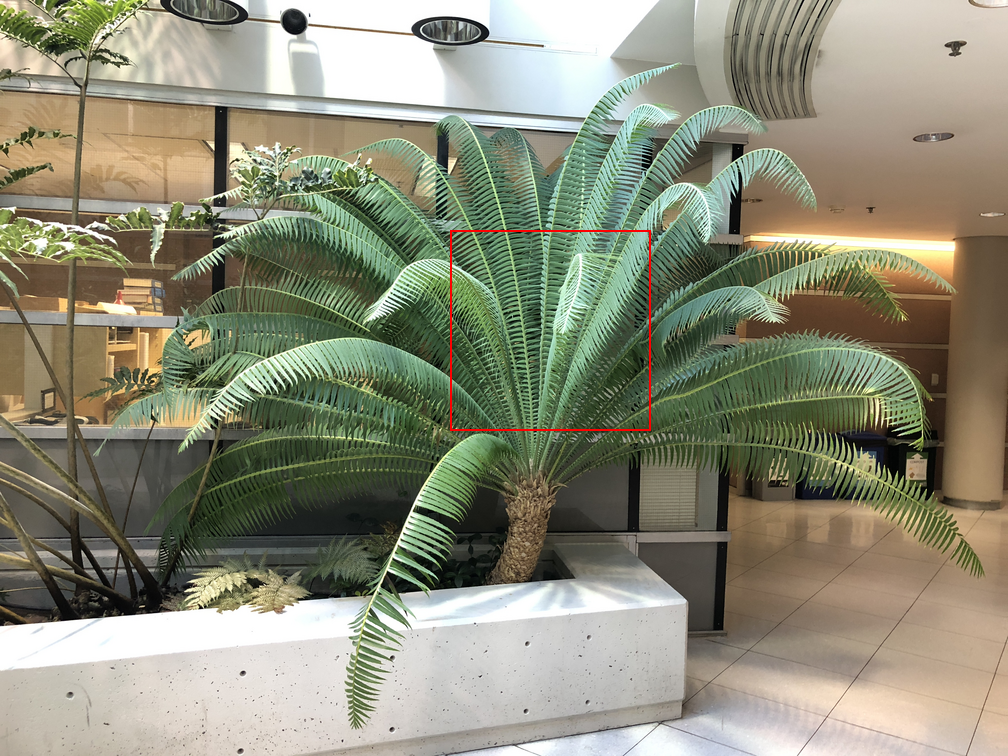} &
\includegraphics[width=0.2\linewidth]{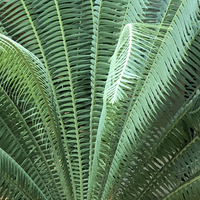} &
\includegraphics[width=0.2\linewidth]{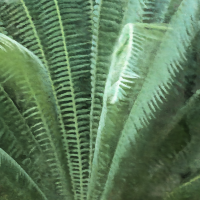} &
\includegraphics[width=0.2\linewidth]{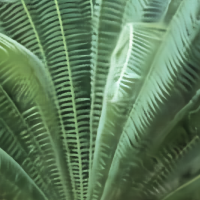} \\
\texttt{Fern} & (a) GT & (b) NeRF & (c) Ours \\
\end{tabular}}
\caption{Visual comparison between our method and NeRF~\cite{nerf} (trained via NeRF-Pytorch~\cite{lin2020nerfpytorch}) on the synthetic $360^{\circ}$ \texttt{Lego} (size: 800$\times$800$\times$3) and real-world forward-facing scene \texttt{Fern} (size: 1008$\times$756$\times$3). \textit{Best viewed in color}. Please refer to our webpage for more visual comparison results.}
\label{fig:visual_comparison}
\end{figure}

\input{figs/2_comp_w_mobilenerf}

\paragraph{Rendering Performance.}
To understand the image quality of various methods, we report three common metrics: PSNR, SSIM~\cite{ssim}, and LPIPS~\cite{lpips}, on the realistic synthetic $360^{\circ}$ and real forward-facing datasets, as demonstrated in Tab.~\ref{table:quantitative}.
Compared with NeRF~\cite{nerf}, our approach achieves better results on PSNR, SSIM, and LPIPS
for the synthetic $360^{\circ}$ dataset. On the forward-facing dataset, we obtain better SSIM and LPIPS than NeRF~\cite{nerf}. Similarly, our method achieves better results for all three metrics than MobileNeRF~\cite{mobilenerf} on the synthetic $360^{\circ}$ dataset and better PSNR and SSIM on the forward-facing dataset.
Compared to SNeRG~\cite{hedman2021baking}, our method obtains better PSNR and SSIM. 

We also show the performance of the teacher model used in training \modelname~in Tab.~\ref{table:quantitative} (Our Teacher). Note that there is still a performance gap between the student model (MobileR2L) and the teacher model. However, as we show following (Tab.~\ref{table:ablation_2}), a better teacher model can lead to a student model with higher performance. Compared with MoibleNeRF and SNeRG, our approach has the advantage that we can directly leverage the high-performing teacher models to help improve student training in different application scenarios. 
We further show the qualitative comparison results in Fig.~\ref{fig:visual_comparison}. On the synthetic scene \texttt{Lego}, our MobileR2L outperforms NeRF clearly, delivering sharper and less distorted shapes and textures. On the real-world scene \texttt{Fern}, our result is less noisy, and the details, \eg, the leaf tips, are sharper. Additionally, we provide the zoom-in comparison with MoibleNeRF~\cite{mobilenerf} in Fig.~\ref{fig:mobile_nerf_comparison}. Our method achieves high-quality rendering for zoom-in view, which is especially important for 3D assets that users might perform zoom-in to look for more image details.


\noindent\textbf{Disk Storage.}
One significant advantage of our method is that we do not require extra storage, even for complex scenes. As shown in Tab.~\ref{table:test_space}, the storage of our approach is $8.3$MB for both synthetic $360^{\circ}$ and forward-facing datasets. The mesh-based methods like MobileNeRF demand more storage for real-world scenes due to saving more complex textures. As a result, our approach takes $24.3\times$ less disk storage than MobileNeRF on the forward-facing, and $15.2\times$ less storage on the synthetic $360^{\circ}$ dataset. 

 
\input{tabs/resources}

\noindent\textbf{Inference Speed.}
We profile and report the rendering speed of our proposed approach on iPhones (13, and 14, iOS 16) in Tab.~\ref{table:speed}. The models are compiled with CoreMLTools~\cite{coreml2021}. Our proposed method runs faster on real forward-facing scenes than the realistic synthetic $360^{\circ}$ scenes. The latency discrepancy between the two datasets comes from the different input spatial sizes. MobileNeRF shows a lower latency than our models on the realistic synthetic $360^{\circ}$ but higher on the real-world scenes. Both methods can run in real-time on devices. Note that MobileNeRF cannot render two scenes, \emph{i.e.}, \texttt{Leaves} and \texttt{Orchids}, due to memory issues, as they require complex textures to model the geometry. In contrast, our approach is robust for different scenes.


\input{tabs/speed}

\noindent\textbf{Discussion.} From the comparison of the rendering quality, disk storage, and inference speed, it can be seen that \modelname~achieves overall better performance than MobileNeRF. More importantly, considering the usage of neural rendering on real-world applications, \modelname~is more suitable as it requires much less storage, reducing the constraint for hardware and can render real-world scenes in real-time on mobile devices.

\subsection{Ablation Analysis}\label{sec:abltation}
Here we perform the ablation analysis to understand the design choices of the network. We use the scene of \texttt{Chair} from the synthetic $360^{\circ}$ to conduct the analysis. All models are trained for $200$K iterations.


\noindent \textbf{Options for Backbone.}
We study the two available operators for designing the backbone.
MLP and $1\times1$ CONV layer are essentially equivalent operators and perform the same calculations, thus resulting in similar performance. However, we observe around $27\%$ latency reduction on mobile devices (iPhone 13) when replacing the MLP layer with the $1\times1$ CONV layer. Specifically, as shown in Tab.~\ref{table:ablation_1}, we design two networks, \emph{i.e.}, MLP and CONV2D, with only residual blocks as in Fig.~\ref{fig:architecture} but removing the activation, normalization, and super-resolution modules. We use the input size as $100\times100$ for the two models. Since the super-resolution modules are omitted, we train the two networks for generating $100\times100$ images.
As can be seen, the CONV2D model has a faster inference speed than MLP with similar performance.
This is due to the fact that CONV operation is better-optimized than MLP on mobile devices. Additionally, due to the intrinsic design of our proposed \modelname, employing MLP layers requires two additional operators, \emph{i.e.}, \texttt{Permute} and \texttt{Reshape}, before feeding the internal features to super-resolution modules, while \texttt{Permute} and \texttt{Reshape} involve data movement that adds unnecessary overheads on some edge devices~\cite{li2022efficientformer}.

\noindent\textbf{Analysis of Activation Function.}
R2L~\cite{r2l} and NeRF~\cite{nerf} use ReLU~\cite{relu} activation as non-linearity function. In our proposed \modelname, we use GELU~\cite{hendrycks2016gaussian} instead. As shown in Tab.~\ref{table:ablation_1}, by comparing the \emph{CONV2D + ReLU} and \emph{CONV2D + GeLU}, which are two networks trained with ReLU and GeLU activations, we notice that GeLU brings about $0.17$ PSNR boost without any additional latency overhead. Similarly, we show that incorporating BatchNorm~\cite{bn} layer into the ResBlock is also beneficial for better performance without introducing extra latency, as shown by \emph{CONV2D + GeLU + BN} in Tab.~\ref{table:ablation_1}. The three networks in the experiments are also trained to render $100\times100$ images.

\noindent\textbf{Analysis on Input Dimension.} We further analyze the optimal spatial resolution for the input tensor. Specifically, we benchmark the performance of three approaches, namely, \emph{$50\times50$ - NeRF Teacher}, \emph{$100\times100$ - NeRF Teacher}, and \emph{$200\times200$ - NeRF Teacher} with the spatial size of input as the square of $50$, $100$, and $200$ respectively. These models contain super-resolution modules to render $800\times800$ images and are trained with the NeRF~\cite{nerf} as a teacher model. Results are presented in Tab.~\ref{table:ablation_2}. The model using a small input spatial size, \emph{i.e.}, $50\times50$, achieves $2\times$ speedup than the model with a larger size, \emph{i.e.}, $100\times100$, as less computation is required. However, the performance is also degraded by $0.25$ PSNR. Further increasing the input spatial size to $200\times200$ makes the model unable to achieve real-time inference. Thus, we do not report the rendering performance of the models with $200\times200$ input size.

\noindent\textbf{Analysis of SR modules.} We further show the necessity of using SR modules. We use the spatial size of $800\times800$ as the network input to render images with the same spatial size. We denote the setting as \emph{$800\times800$ - w/o SR} in Tab.~\ref{table:ablation_2}. Profiling the latency of such a network leads to compilation errors due to intensive memory usage. Thus, our proposed SR module is \textit{essential} for high-resolution synthesis without introducing prohibitive computation overhead.

\noindent\textbf{Choice of Teacher Models.} 
Since both R2L~\cite{r2l} and \modelname~ use a teacher model for generating pseudo data to train a lightweight network, we study whether a more powerful teacher model can improve performance. To conduct the experiments, we use MipNeRF~\cite{mip-nerf} as the teacher model for the training \modelname, given MipNeRF shows better performance than NeRF on the synthetic $360^{\circ}$ dataset. We denote the approach as \emph{$100\times100$ - MipNeRF Teacher}. Through the comparison with \emph{$100\times100$ - NeRF Teacher}, as in Tab.~\ref{table:ablation_2}, we notice the quality of the rendered images is improved, \emph{e.g.}, the PSNR is increased by $0.18$, without the extra cost of latency. The comparison demonstrates that our approach has the potential to render higher-quality images when better teacher models are provided.

\input{tabs/ablation}

\noindent\textbf{Analysis on Ray Representation.}
Here we analyze how the ray representation affects the latency-performance trade-off of \modelname.
We follow the same ray representation paradigm as in R2L~\cite{r2l} and positional encoding as in NeRF~\cite{nerf}. Specifically, R2L sample $K$ 3D points along the ray, and each point is mapped to a higher dimension by positional encoding with $L$ positional coordinates. R2L sets $K = 16$ and $L=10$, resulting in a vector with a dimension of $1,008$. We apply the same setting to \modelname~and denote the model as \emph{$100\times100$, $K16$, $L10$ - MipNeRF Teacher} in Tab.~\ref{table:ablation_2}. For our implementation in \modelname, we set $K=8~\text{and}~L=6$ for the model, which has the dimension per ray as $312$. The model is \emph{$100\times100$ - MipNeRF Teacher}. By comparing the performance of the two models, we notice that larger $K$ and $L$ lead to negligible PSNR improvement ($0.07$), yet higher inference latency and bigger model size (more parameters).
Therefore, we chose $K=8~\text{and}~L=6$ for the consideration of model size and latency -- they are the more important  metrics when deploying neural rendering models on mobile devices.


\noindent\textbf{Depth of the Backbone.} Lastly, we show the effects of the backbone depth on the model performance. We use the depth as $60$ for our backbone. By increasing the number of residual blocks in the backbone, \emph{i.e.}, setting depth as $100$, we obverse better model performance at the cost of higher latency, as shown by comparing the last two rows in Tab.~\ref{table:ablation_2}. Using depth as $100$ significantly increases the latency and the number of parameters, and the model fails to run in real time. Thus, we chose depth as $60$, given the better trade-off between latency, model size, and performance.

\subsection{Real-World Application}
Here we demonstrate the usage of our technique for building a real-world application. Given the small size and faster inference speed of our model, we create a shoes try-on application that runs on mobile devices. Users can directly try on the shoes rendered by \modelname~using their devices, enabling real-time interaction. 

The pipeline for building the application is illustrated in Fig.~\ref{fig:try_on}. We first use iPhone to capture $100$ shoe images for training. The images are then segmented to remove the background. After that, we train a NeRF model~\cite{nerf} to generate pseudo data, which is later used for learning \modelname~to render images in $1008\times756$ resolution. We apply foot tracking and overlay the rendered shoe on top of the user's feet. As can be seen from Fig.~\ref{fig:try_on}, our model is able to render high-quality images from various views for different users. The try-on usage proves the potential of leveraging neural rendering for building various real-time interactive applications such as Augment Reality.

\input{figs/3_try_on_demo}

%% file: tabs/quantitative_comparison.tex
\begin{table}
\begin{center}
\caption{\textbf{Quantitative Comparison} on Synthetic $360^{\circ}$ and Forward-facing. Our method obtains better results on the three metrics than NeRF for the two datasets. Compared with MoibleNeRF and SNeRG, we achieve better results on most of the metrics.}\label{table:quantitative}
\resizebox{1.0\linewidth}{!}{
\setlength{\tabcolsep}{2pt}
\begin{tabular}{l|ccc|ccc}
\hline
 & \multicolumn{3}{c|}{Synthetic $360^{\circ}$} & \multicolumn{3}{c}{Forward-facing} \\
 & PSNR$\uparrow$ & SSIM$\uparrow$ & LPIPS$\downarrow$ & PSNR$\uparrow$ & SSIM$\uparrow$ & LPIPS$\downarrow$  \\
\hline
NeRF~\cite{nerf} & 31.01 & 0.947 & 0.081 & 26.50 & 0.811 & 0.250  \\
NeRF-Pytorch~\cite{lin2020nerfpytorch} & 30.92 & 0.991 & 0.045 & 26.26 & 0.965 & 0.153  \\
\hline
SNeRG~\cite{hedman2021baking} & 30.38 & {0.950} & \textbf{0.050} & 25.63 & 0.818 & \textbf{0.183}  \\
MoibleNeRF~\cite{mobilenerf} & {30.90} & 0.947 & 0.062 & {25.91} & {0.825} & \textbf{0.183}   \\
{MobileR2L (Ours)}  & \textbf{31.34} & \textbf{0.993} & 0.051 &\textbf{26.15}  & \textbf{0.966} &{0.187}  \\\hline
{Our Teacher}  & 33.09 &  0.961 & 0.052 & 26.85  & 0.827 &  0.226 \\

\hline
\end{tabular}
}

\end{center}
\end{table}

%% file: figs/2_comp_w_mobilenerf.tex
\begin{figure}[ht]
  \includegraphics[width=1\linewidth]{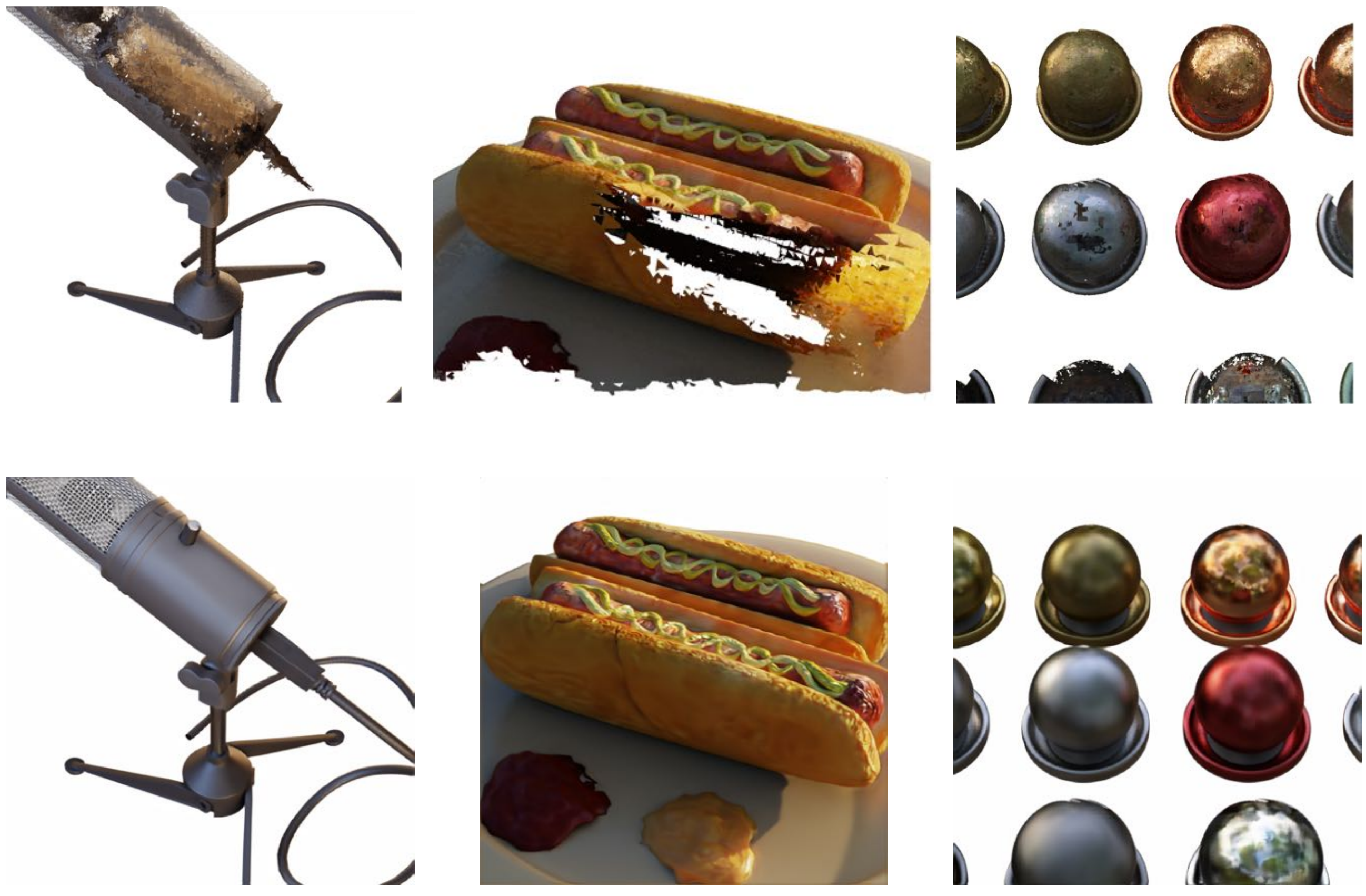}  
  \caption{Zoom-in comparisons. \emph{Top row}: MobileNeRF~\cite{mobilenerf}. Results are obtained from the code and demo released by the authors. \emph{Bottom row}: \modelname. Our approach renders high-quality images even for zoom-in views.}
  \label{fig:mobile_nerf_comparison}
\end{figure}

%% file: tabs/resources.tex
\begin{table}
\begin{center}
\caption{\textbf{Analysis of Storage (MB)} required for different rendering methods. Our method has a clear advantage over existing works with much less storage required.}\label{table:test_space}
\resizebox{1.0\linewidth}{!}{
\begin{tabular}{l|ccc|ccc}
\hline
 & \multicolumn{3}{c|}{Synthetic $360^{\circ}$} & \multicolumn{3}{c}{Forward-facing}  \\
 & MoibleNeRF~\cite{mobilenerf} & SNeRG~\cite{hedman2021baking} & {Ours} & MobileNeRF~\cite{mobilenerf} & SNeRG~\cite{hedman2021baking} & {Ours} \\
\hline
Disk storage & 125.8 & {86.8} & \textbf{8.3}& {201.5} & 337.3 & \textbf{8.3}  \\
\hline
\end{tabular}
}
\end{center}
\end{table}

%% file: tabs/speed.tex
\begin{table}[h]
\begin{center}
\caption{
\textbf{Analysis of Inference Speed.}
Latency (ms) is obtained on iPhone with iOS 16. Follwing MoibleNeRF~
\cite{mobilenerf}, we use the notation $\frac{M}{N}$ to indicate that $M$ out of $N$ scenes in the Forward-facing dataset that can not run on devices. Specifically, MobileNeRF can not render \texttt{Leaves} and \texttt{Orchids} in Forward-facing.}\label{table:speed}
\resizebox{1\linewidth}{!}{
\begin{tabular}{l|cc|cc}
\hline
 & \multicolumn{2}{c|}{Synthetic $360^{\circ}$} & \multicolumn{2}{c}{Forward-facing}  \\
 & MobileNeRF~\cite{mobilenerf} & {Ours} & MobileNeRF~\cite{mobilenerf} & {Ours} \\
\hline
{iPhone 13} & 17.54 & 26.21 & {27.15}\footnotesize{$\frac{2}{8}$} & 18.04      \\
{iPhone 14} & 16.67 & 22.65 & {20.98}\footnotesize{$\frac{2}{8}$} &   16.48    \\
\hline
\end{tabular}
}
\end{center}
\end{table}

%% file: tabs/ablation.tex
\begin{table}[t]
\begin{center}
\caption{
\textbf{Analysis of Network Design.} For all the comparisons, we use the input tensor with the spatial size as $100\times100$ and render the image with spatial size. The latency (ms) is measured on iPhone 13 (iOS16) with models compiled with CoreMLTools~\cite{coreml2021}}\label{table:ablation_1}
\resizebox{1\linewidth}{!}{
\begin{tabular}{l|cccc}
\hline
 & PSNR$\uparrow$ & SSIM$\uparrow$ & LPIPS$\downarrow$ & Latency$\downarrow$ \\
\hline
{MLP} &  19.13 &0.9759  &0.6630 &  19.57 \\
{CONV2D} & 19.16  & 0.9759  & 0.6301 &  14.30 \\
{CONV2D + ReLU} &  26.82 &0.9949  & 0.0282  &  16.20  \\
{CONV2D + GeLU } &26.99  & 0.9949 &0.0730 &  17.00 \\ 
{CONV2D + GeLU + BN} & \textbf{27.18} & \textbf{0.9954}  & \textbf{0.0259}  &  17.00\\
\hline
\end{tabular}
}
\end{center}
\end{table}

\begin{table}[t]
\begin{center}
\caption{
\textbf{Analysis of the spatial size of the input, usage of teacher model, and ray presentation}. Besides image quality metrics, we show the number of parameters for each model and the latency when running on iPhone 13. }\label{table:ablation_2}
\resizebox{1\linewidth}{!}{
\setlength{\tabcolsep}{1mm}
\begin{tabular}{lccccc}
\toprule
 & Params & PSNR$\uparrow$ & SSIM$\uparrow$ & LPIPS$\downarrow$ & Latency$\downarrow$ \\
\midrule
{$50\times50$ - NeRF Teacher} &3.9M &30.40 & 0.9965 & 0.0686 &  13.04 \\
{$100\times100$ - NeRF Teacher} &3.9M & 30.65 &0.9966  &0.0668  & 26.21 \\
{$200\times200$ - NeRF Teacher} &3.9M & - &- &-  &  73.76 \\
{$800\times800$ - w/o SR} &3.9M & - & - & - &  Error \\ \hdashline
{$100\times100$ - MipNeRF Teacher} & 3.9M&  30.83 & 0.0997 & 0.0564 & 26.21 \\ 
{$100\times100$ - MipNeRF Teacher, $K16$, $L10$} &4.1M & 30.90 &0.9968  & 0.0583  & 31.05  \\ 
$100\times100$ - MipNeRF Teacher, $K16$, $L10$, D100 &6.8M & 31.37 & 0.9972 &0.0470  & 44.52 \\
\bottomrule
\end{tabular}
}
\end{center}
\end{table}

%% file: figs/3_try_on_demo.tex
\begin{figure}[ht]
  \includegraphics[width=1.0\linewidth]{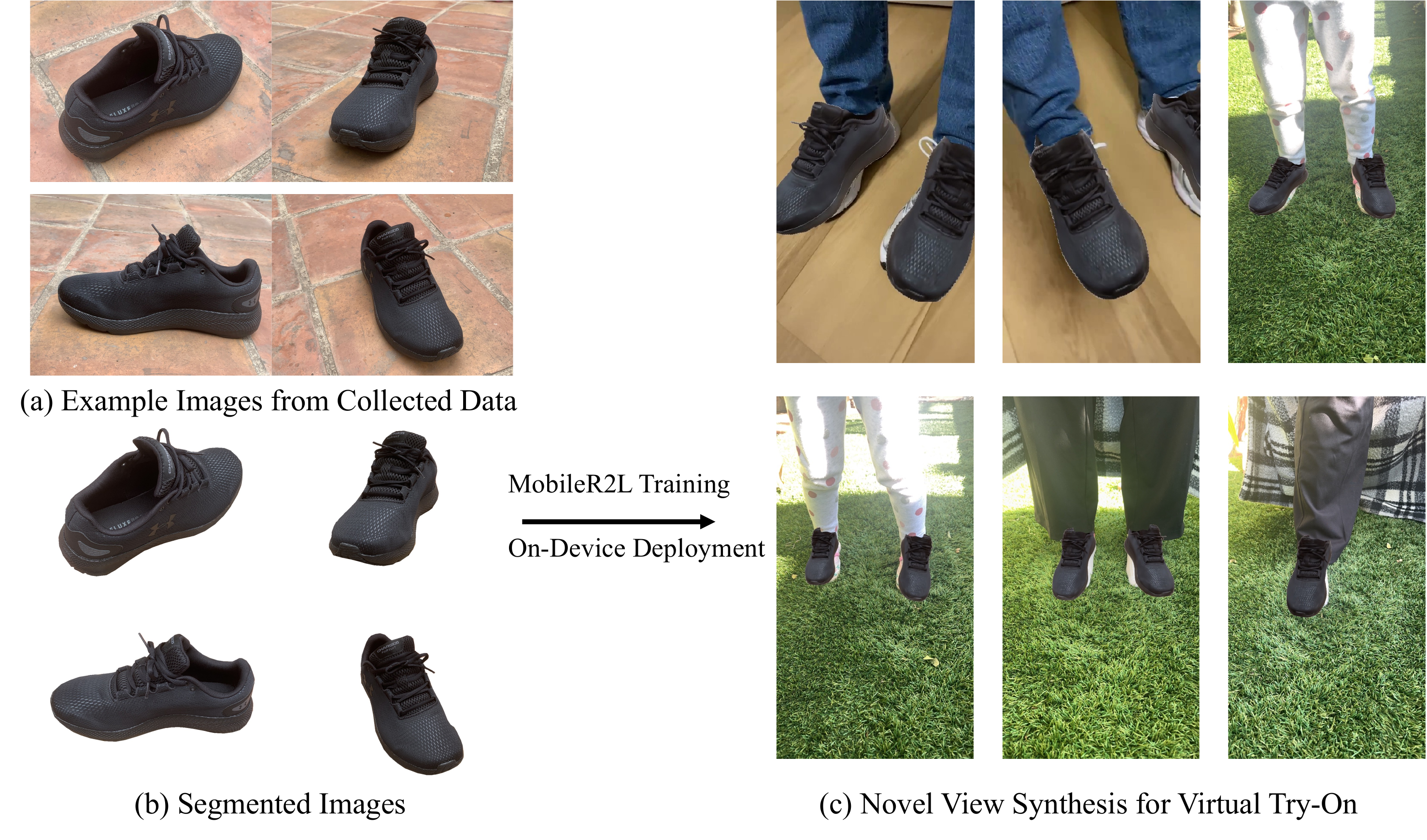}  
  \caption{\textbf{Virtual Try-On Application.} From the collected images using a cellphone (a), we segment the foreground shoe (b) to train a \modelname~model. We deploy the model on mobile devices, and users can directly try the shoe. The model renders images for novel views when users rotate the phone or change the foot positions (c).}
  \label{fig:try_on}
\end{figure}

%% file: section/5_conclusion.tex
\begin{figure}[t]
\centering
\resizebox{1\linewidth}{!}{
\setlength{\tabcolsep}{0.5mm}
\begin{tabular}{ccccc}
\includegraphics[width=0.267\linewidth]{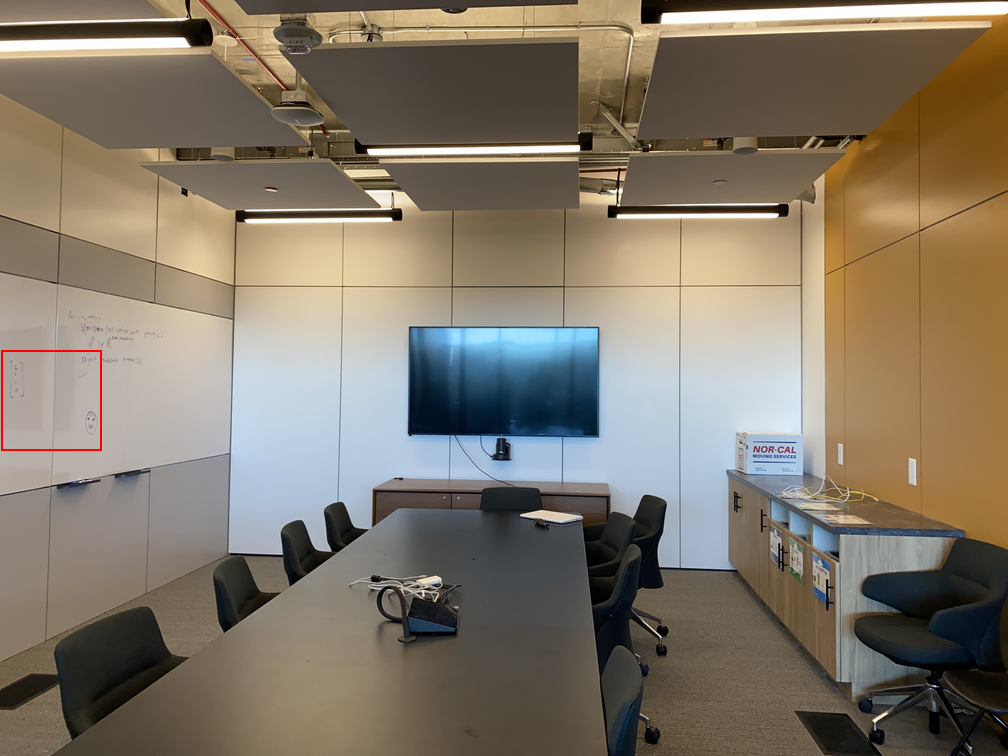} &
\includegraphics[width=0.2\linewidth]{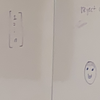} &
\includegraphics[width=0.2\linewidth]{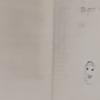} &
\includegraphics[width=0.2\linewidth]{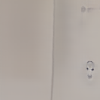} \\
\texttt{Room} & (a) GT & (b) NeRF & (c) Ours \\
\end{tabular}}
\caption{Visual comparison on the real-world scene \texttt{Room}. Both our model and NeRF fail to synthesize the whiteboard writings on the upper-left of the cutout patch.}
\label{fig:limitation}
\vspace{-2mm}
\end{figure}

\section{Limitation and Conclusion}
This work presents \modelname, the first neural light network that renders images with similar quality as NeRF~\cite{nerf} while running in real-time on model devices. We perform extensive experiments to design an optimized network architecture that can be trained via data distillation to render high-resolution images, \emph{e.g.}, $800\times800$. Additionally, since we do not require other information besides the neural network, \modelname~dramatically saves the representation storage in stark contrast to  other mesh-based methods like MoibleNeRF~\cite{mobilenerf}. Furthermore, we prove that with our design, neural rendering can be used to build real-world applications, achieving real-time user interaction.

Although \modelname~achieves promising inference speed with small model sizes, there are still two limitations of the current work that can be improved. First, we follow the training recipe of R2L~\cite{r2l}, and R2L uses $10$K pseudo images generated by the teacher NeRF model to train the student model. The number of training images is much more than the images used to train the teacher NeRF (which only requires around $100$ images), resulting in a longer training time than NeRF-based methods. Therefore, a future direction could be reducing the training costs for distillation-based works like R2L and this work. Second, \modelname~fails to generate some high-frequency details in the images. We show examples in Fig.~\ref{fig:limitation}. Using a larger model may alleviate this problem given a larger model capacity. Nevertheless, the inference latency will also increase accordingly on mobile devices. Future efforts may focus on optimizing the network and training pipeline to boost the performance of the current model.

%% file: section/6_supp.tex
\section{Additional Ablation Analysis}
Besides the ablation study shown in Sec.~\ref{sec:abltation}, we provide more analyses of choices when we design the network architecture. The baseline network has the backbone with depth as $60$ layers and $3$ Residual Blocks (RB) in each Super-Resolution (SR) module. We use the data from \texttt{Chair} in the synthetic 360$^{\circ}$ dataset to train all models with $300$K iterations on Nvidia V100 GPU with batch size as $8$. The latency is profiled using iPhone 13 (iOS 16) with CoreMLTools~\cite{coreml2021}. For the models that are too big to fit into the Nvidia V100 GPU, we only benchmark the latency instead of training the models.
The conducted comparisons are introduced as follows:
\setlist{nosep}
\begin{itemize}[leftmargin=1em]
    \item \textbf{Number of RB per SR module.} We verify the number of RB in each SR module. From Tab.~\ref{table:ablation}, using $3$ RB in each SR model, \emph{i.e.}, 3RB per SR, achieves a better trade-off between network performance, \emph{e.g.}, PSNR, and latency.
    \item \textbf{Number of output channels in SR.} We change the number of output channels of the RB in SR. We choose the design of $64$ output channels in the first two SR modules and $16$ channels in the last SR module, \emph{i.e.}, C64-64-16 in Tab.~\ref{table:ablation}, because it has a real-time inference speed on the mobile device while maintaining a good performance.
    \item \textbf{Width of backbone.} We adopt different width, \emph{i.e.,} the number of channels in the convolution layers, for the backbone. The network with the width as $256$, \emph{i.e.,} W256 in Tab.~\ref{table:ablation}, gives us the better trade-off between latency and network performance.
    \item \textbf{Depth of backbone.} Lastly, we show how the depth, \emph{i.e.,} the number of layers, in the backbone affects the performance. We chose the depth as $60$, \emph{i.e.,} D60 in Tab.~\ref{table:ablation}, due to the better PSNR and satisfied latency.
\end{itemize}


\section{Per-Scene Quantitative Results} \label{sec:per-scene_results}
Here we provide detailed per-scene comparison results (PSNR, SSIM, and LPIPS) on the synthetic 360$^{\circ}$ (Tab.~\ref{tab:per-scene-blender-psnr}, Tab.~\ref{tab:per-scene-blender-ssim}, and Tab.~\ref{tab:per-scene-blender-lpips}) and forward-facing (Tab.~\ref{tab:per-llff-psnr}, Tab.~\ref{tab:per-llff-ssim}, and Tab.~\ref{tab:per-llff-lpips}) datasets. Please note that as we implement our framework following NeRF-Pytorch~\cite{lin2020nerfpytorch} and R2L~\cite{r2l}, we use the same evaluating pipeline for measuring the quantitative metrics. 
We also compare our results with the models trained by NeRF-Pytorch~\cite{lin2020nerfpytorch}. As can be seen from the comparisons, our approach (MobileR2L) achieves comparable or even better results than NeRF~\cite{nerf} and NeRF-Pytorch~\cite{lin2020nerfpytorch}. 

\section{More Visual Comparisons} \label{sec:more_visual_results}
In Fig.~\ref{fig:more_visual_comparison}, we present more visual comparison results on the synthetic 360$^{\circ}$ and forward-facing datasets. We note, (1) MobileR2L can produce the \textit{correct} texture details that NeRF cannot, \eg, on the scene \texttt{Mic}, NeRF almost loses the grid texture of the Mic while our MobileR2L manages to render it out; similarly, on the scene \texttt{Horn}, there is (at least) one button on the glass wall missed by NeRF while our MobileR2L does not. (2) When both  MobileR2L and NeRF can render out the details, MobileR2L typically generates \textit{clearer, sharper, and less noisy} results: on the scene \texttt{T-Rex}, it is obvious that our MobileR2L renders much less noisy railing; similar phenomenon can also be observed on the scene \texttt{Hotdog}, \texttt{Material}, and \texttt{Drum}.


\section{Power Usage} \label{sec: power_usage}

 We profile the power usage of our proposed method on MacBook Pro (chip: Apple M1 Pro, OS: Ventura, V13.1) with the public tool\footnote{https://github.com/tlkh/asitop}, due to no publicly available tools for benchmarking the power usage on iPhones. Tab.~\ref{tab:power-usage} shows the power usage (in W) when running MobileNeRF (on GPU) and our work (on the neural engine) for the Synthetic $360^{\circ}$ and Forward-facing datasets. Our work consumes less power than MobileNeRF, especially for real-world scenes (7.7$\times$ less on Forward-facing). MobileNeRF requires to load complicated textures while ours does not.
We also profile the CPU and RAM usage for MobileNeRF and our work, which are similar (RAM: 1GB, CPU: 2W).

\begin{table}[ht]
\begin{center}
\caption{
\textbf{Ablation analysis on network architectures.} We report the number of parameters (\#Params), PSNR, SSIM, LPIPS, and Latency (ms, on iPhone 13) for each design choice.}\label{table:ablation}
\resizebox{1\linewidth}{!}{
\setlength{\tabcolsep}{1mm}
\begin{tabular}{lccccc}
\toprule
 & \#Params & PSNR$\uparrow$ & SSIM$\uparrow$ & LPIPS$\downarrow$ & Latency$\downarrow$ \\
\midrule
{1RB per SR} &3.9M & 31.58 & 0.9973 & 0.0503 & 22.38 \\
{2RB per SR} &3.9M & 31.63 & 0.9973 &0.0484   & 26.21 \\
{3RB per SR} &3.9M & 31.73 & 0.9973 & 0.0368  &  30.42 \\
{4RB per SR} &4.0M & 31.59 & 0.9972 & 0.0508 &  33.80 \\ \hdashline

{C16-16-16} & 3.8M&  31.01 & 0.9969 & 0.0644 & 22.77 \\ 
{C32-32-32}& 3.9M & 31.39 & 0.9972 &  0.0525  & 35.77  \\
{C64-64-16}& 3.9 M &31.63 & 0.9973 &0.0484  & 26.21 \\ 
{C64-64-64} &3.9M & - & - & -  & 59.31 \\ 
 \hdashline

{W64} & 0.3M&  28.83 & 0.9951 & 0.0896 & 13.22 \\ 
{W128 }&1.0M & 30.23 &0.9963  & 0.0699  & 17.91 \\
{W256} &3.9M & 31.63 & 0.9973 &0.0484  & 26.21\\ 
{W384}& 8.7M & 32.28 &0.9977  &0.0359  & 39.08  \\ 
{W512} &15.4M & - & - & -  & 53.47 \\ \hdashline

{D30} & 1.9M&  30.86 & 0.9965 & 0.0609 & 18.72 \\ 
{D60} & 3.9M & 31.63 & 0.9973 &0.0484  & 26.21\\ 
{D80}& 5.2M & 31.60 & 0.9972 & 0.0499 & 31.49  \\ 
{D100} &6.6M & - & - & -  & 36.55 \\ 
\bottomrule
\end{tabular}
}
\end{center}
\end{table}

\begin{table}
  \centering
  \begin{minipage}[c]{1\linewidth}
        \centering
\caption{Per-scene PSNR$\uparrow$ comparison on the Synthetic 360$^{\circ}$ dataset between NeRF~\cite{nerf}, NeRF-Pytorch~\cite{lin2020nerfpytorch}, and our approach.}
\resizebox{1\linewidth}{!}{
\begin{tabular}{l|c|c|c|c|c|c|c|c|c}
\toprule
Method & Chair &  Drums & Ficus &  Hotdog &  Lego  &  Materials & Mic &  Ship  & Average\\\hline 
NeRF~\cite{nerf} & 
33.00 & 25.01 & 30.13 & 36.18 & 32.54 & 29.62 & 32.91 & 28.65 & 31.01\\
NeRF-Pytorch~\cite{lin2020nerfpytorch} & 33.31 & 25.14 & 30.28  & 36.52& 31.80  & 29.25&32.50 & 28.54 & 30.92 \\
MobileR2L (Ours) & 33.66 & 25.05 & 29.80  & 36.84& 32.18  & 30.54&34.37 & 28.75 & 31.34 \\
\bottomrule
\end{tabular}
}\label{tab:per-scene-blender-psnr}
  \end{minipage}
\\
\begin{minipage}[c]{1\linewidth}
\centering
\caption{Per-scene SSIM$\uparrow$ comparison on the Synthetic 360$^{\circ}$ dataset between NeRF~\cite{nerf}, NeRF-Pytorch~\cite{lin2020nerfpytorch}, and our approach.}
\resizebox{1\linewidth}{!}{
\begin{tabular}{l|c|c|c|c|c|c|c|c|c}
\toprule
Method & Chair &  Drums & Ficus &  Hotdog &  Lego  &  Materials & Mic &  Ship  & Average\\\hline 
NeRF~\cite{nerf} & 0.967 &0.925& 0.964& 0.974& 0.961& 0.949& 0.980& 0.856 &0.947
 \\
NeRF-Pytorch~\cite{lin2020nerfpytorch} & 0.998 & 0.985 & 0.996  & 0.998& 0.991  & 0.989&0.996 & 0.980&0.991 \\
MobileR2L (Ours) & 0.998 & 0.986 & 0.996  & 0.998& 0.992  & 0.992&0.997 & 0.982&0.993 \\
\bottomrule
\end{tabular}
}
\label{tab:per-scene-blender-ssim}
  \end{minipage}

\begin{minipage}[c]{1\linewidth}
\centering
\caption{Per-scene LPIPS$\downarrow$ comparison on the Synthetic 360$^{\circ}$ dataset between NeRF~\cite{nerf}, NeRF-Pytorch~\cite{lin2020nerfpytorch}, and our approach.}
\resizebox{1\linewidth}{!}{
\begin{tabular}{l|c|c|c|c|c|c|c|c|c}
\toprule
Method & Chair &  Drums & Ficus &  Hotdog &  Lego  &  Materials & Mic &  Ship  & Average\\\hline 
NeRF~\cite{nerf} & 0.046 &0.091& 0.044& 0.121& 0.050& 0.063& 0.028& 0.206 &  0.081 \\
NeRF-Pytorch~\cite{lin2020nerfpytorch} & 0.025 & 0.066 & 0.023  & 0.022& 0.029  & 0.035&0.021 & 0.144& 0.045 \\
MobileR2L (Ours) & 0.027 & 0.083 & 0.025  & 0.026& 0.043  & 0.029&0.012 & 0.162&0.051 \\
\bottomrule
\end{tabular}
}
\label{tab:per-scene-blender-lpips}
  \end{minipage}

 \begin{minipage}[c]{1\linewidth}
\centering
\caption{Per-scene PSNR$\uparrow$ comparison on the  Forward-facing dataset between NeRF~\cite{nerf}, NeRF-Pytorch~\cite{lin2020nerfpytorch}, and our approach.}
\resizebox{1\linewidth}{!}{
\begin{tabular}{l|c|c|c|c|c|c|c|c|c}
\toprule
Method & Room & Fern & Leaves  &Fortress & Orchids & Flower & T-Rex& Horns & Average\\\hline 
NeRF~\cite{nerf} & 32.70 & 25.17&  20.92&  31.16&  20.36&  27.40&  26.80&  27.45& 26.50\\
NeRF-Pytorch~\cite{lin2020nerfpytorch} & 32.10 & 24.80 & 20.50  & 31.20& 20.45  & 27.50 &26.48 & 27.05&26.26 \\
MobileR2L (Ours) & 32.09 & 24.39 & 20.52  & 30.81& 20.06  & 27.61&26.71 & 27.01&26.15  \\
\bottomrule
\end{tabular}
}
\label{tab:per-llff-psnr}
  \end{minipage}
\hfill
\begin{minipage}[c]{1\linewidth}
\centering
\caption{Per-scene SSIM$\uparrow$ comparison on the  Forward-facing dataset between NeRF~\cite{nerf}, NeRF-Pytorch~\cite{lin2020nerfpytorch}, and our approach.}
\resizebox{1\linewidth}{!}{
\begin{tabular}{l|c|c|c|c|c|c|c|c|c}
\toprule
Method & Room & Fern & Leaves  &Fortress & Orchids & Flower & T-Rex& Horns & Average\\\hline 
NeRF~\cite{nerf} &0.948  &0.792 & 0.690  &0.881 & 0.641 & 0.827 & 0.880 & 0.828 &0.811\\
NeRF-Pytorch~\cite{lin2020nerfpytorch} & 0.989 & 0.976 & 0.921  & 0.995& 0.920  & 0.968&0.972 & 0.983&0.965 \\
MobileR2L (Ours) & 0.995 & 0.973 & 0.923  & 0.995& 0.916  &0.971 & 0.973 & 0.982&0.966 \\
\bottomrule
\end{tabular}
}
\label{tab:per-llff-ssim}
  \end{minipage}

\begin{minipage}[l]{1\linewidth}

\caption{Per-scene LPIPS$\downarrow$ comparison on the Forward-facing dataset between NeRF~\cite{nerf}, NeRF-Pytorch~\cite{lin2020nerfpytorch}, and our approach.}
\resizebox{1\linewidth}{!}{
\begin{tabular}{l|c|c|c|c|c|c|c|c|c}
\toprule
Method & Room & Fern & Leaves  &Fortress & Orchids & Flower & T-Rex& Horns & Average\\\hline 
NeRF~\cite{nerf} & 0.178 &0.280& 0.316& 0.171 &0.321& 0.219& 0.249& 0.268&0.250\\
NeRF-Pytorch~\cite{lin2020nerfpytorch} & 0.089 & 0.210 & 0.921  & 0.995& 0.920  & 0.968&0.972 & 0.983& 0.153 \\
MobileR2L (Ours) & 0.088 & 0.239 & 0.280  & 0.103 & 0.296  & 0.150 & 0.121 & 0.217 &0.187 \\
\bottomrule
\end{tabular}
}
\label{tab:per-llff-lpips}
  \end{minipage}

\begin{minipage}[l]{1\linewidth}
\caption{Power usage on  the Synthetic 360$^{\circ}$ dataset and  the  Forward-facing dataset between MobileNeRF\cite{mobilenerf} and our approach.}
\resizebox{1\linewidth}{!}{
\begin{tabular}{lccccccccc}

\hline 
Synthetic $360^{\circ}$ & Chair & Drums & Ficus & Hotdog & Lego & Material & Mic & Ship & Avg$\downarrow$\\
\hline
{MobileNeRF}& 1.7W & 1.6W & 1.4W & 4.3W & 2.6W & 2.1W & 1.2W & 7.3W & 2.8W \\
\rowcolor[gray]{0.92}
{Ours} & 2.5W &2.5W &2.5W &2.5W &2.5W &2.5W &2.5W &2.5W &2.5W \\
\hline\hline
Forward-facing& Fern & Flower & Fortress & Horn & Leaves & Orchids & Room & Trex & Avg$\downarrow$\\
\midrule
{MobileNeRF} & 12.3W & 13.0W & 12.4W & 12.8W & 15.1W & 14.5W & 12.8W & 12.9W & 13.2W\\
\rowcolor[gray]{0.92}
{Ours} & 1.7W &1.7W &1.7W &1.7W &1.7W &1.7W &1.7W &1.7W &1.7W   \\
\hline
\end{tabular}

}
\label{tab:power-usage}
    \end{minipage}

\end{table}

\begin{figure}[ht]
\centering
\resizebox{1\linewidth}{!}{
\setlength{\tabcolsep}{0.5mm}
\begin{tabular}{ccccc}
\includegraphics[width=0.2\linewidth]{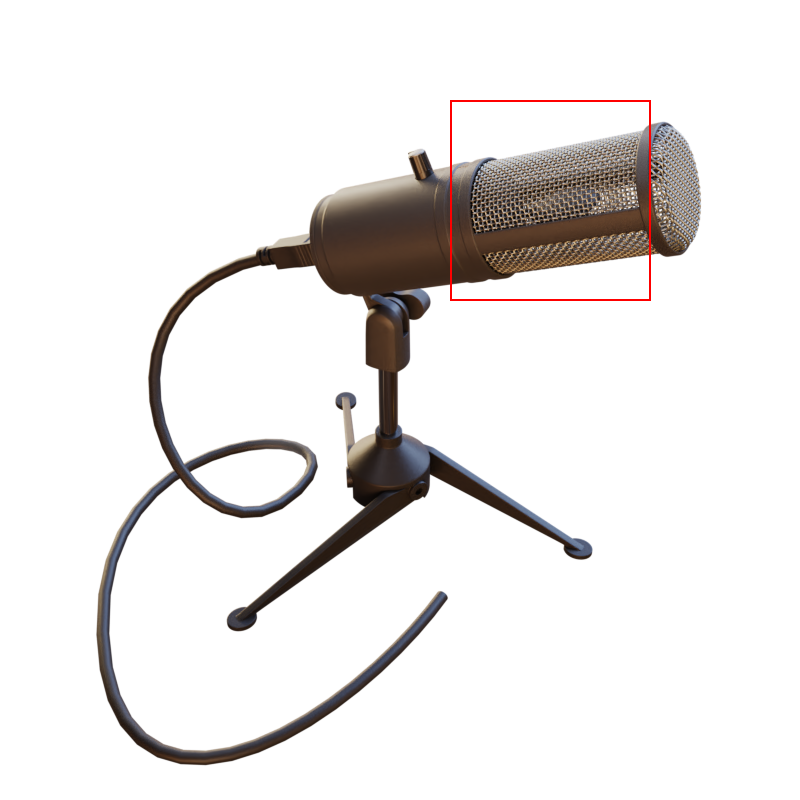} &
\includegraphics[width=0.2\linewidth]{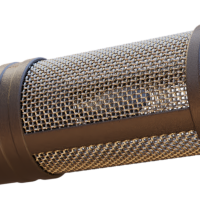} &
\includegraphics[width=0.2\linewidth]{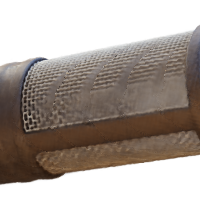} &
\includegraphics[width=0.2\linewidth]{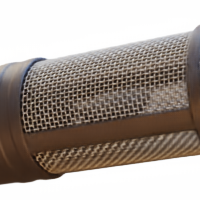} \\
\texttt{Mic} & (a) GT & (b) NeRF & (c) Ours \\
\includegraphics[width=0.2\linewidth]{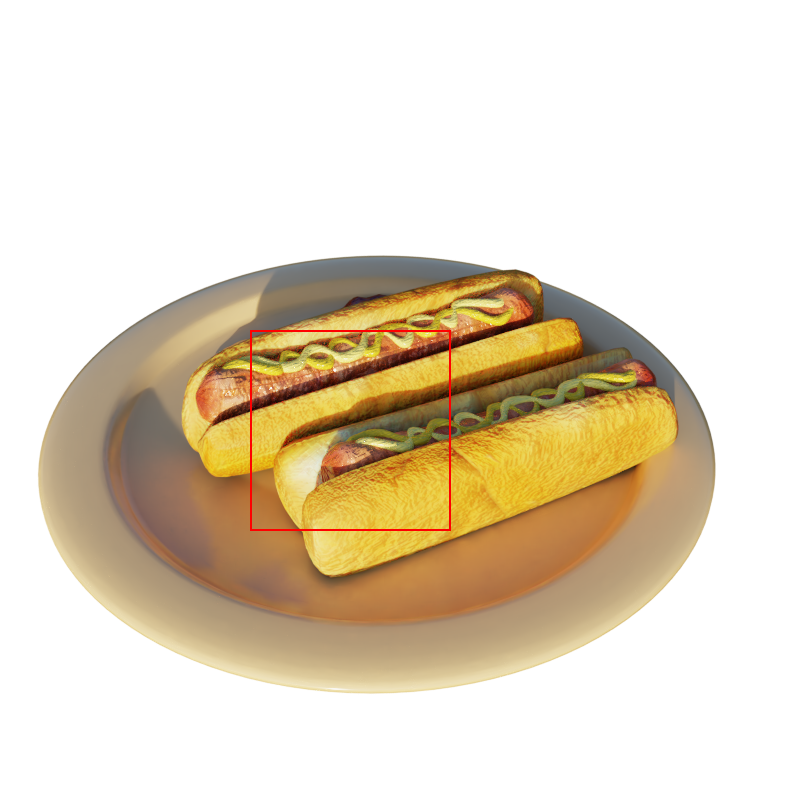} &
\includegraphics[width=0.2\linewidth]{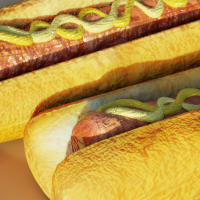} &
\includegraphics[width=0.2\linewidth]{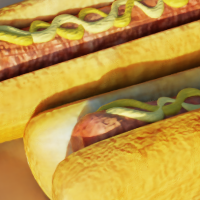} &
\includegraphics[width=0.2\linewidth]{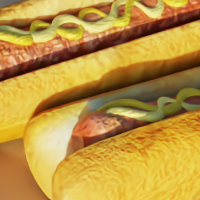} \\
\texttt{Hotdog} & (a) GT & (b) NeRF & (c) Ours \\
\includegraphics[width=0.2\linewidth]{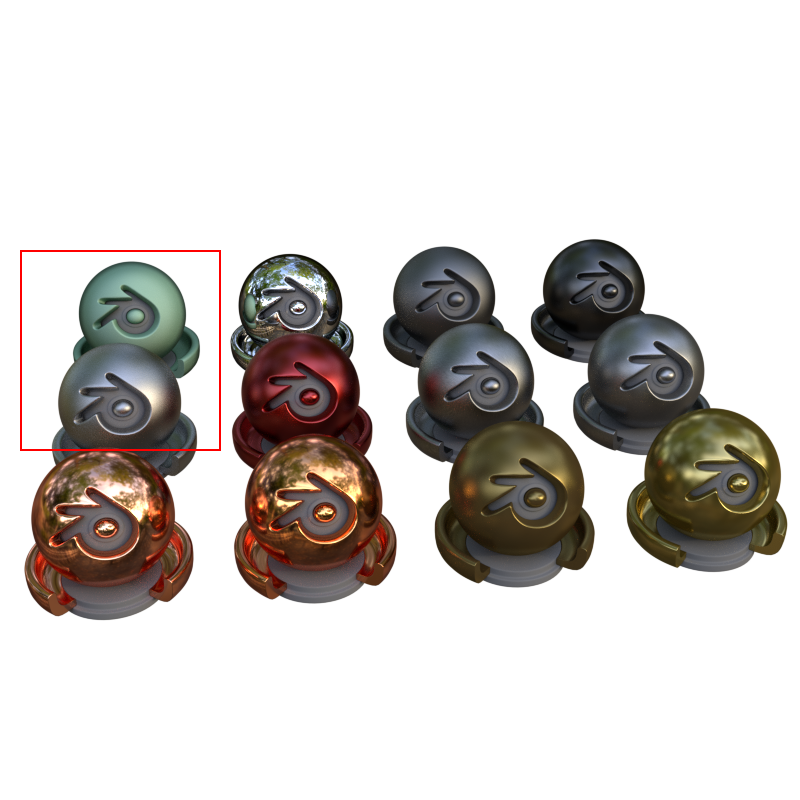} &
\includegraphics[width=0.2\linewidth]{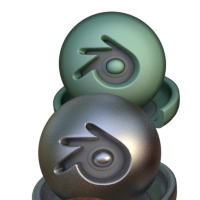} &
\includegraphics[width=0.2\linewidth]{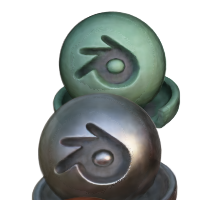} &
\includegraphics[width=0.2\linewidth]{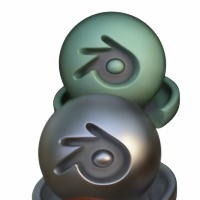} \\
\texttt{Material} & (a) GT & (b) NeRF & (c) Ours \\
\includegraphics[width=0.2\linewidth]{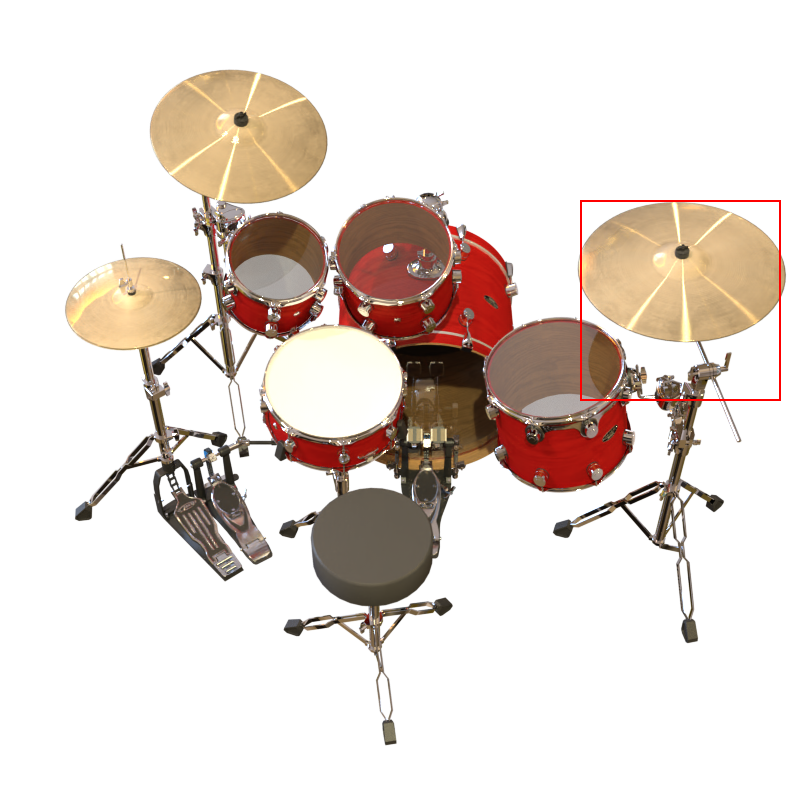} &
\includegraphics[width=0.2\linewidth]{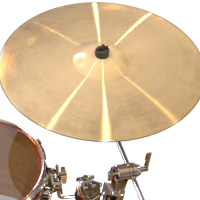} &
\includegraphics[width=0.2\linewidth]{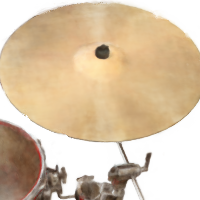} &
\includegraphics[width=0.2\linewidth]{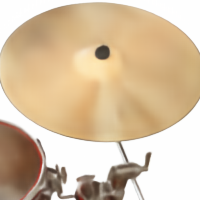} \\
\texttt{Drum} & (a) GT & (b) NeRF & (c) Ours \\
\includegraphics[width=0.2\linewidth]{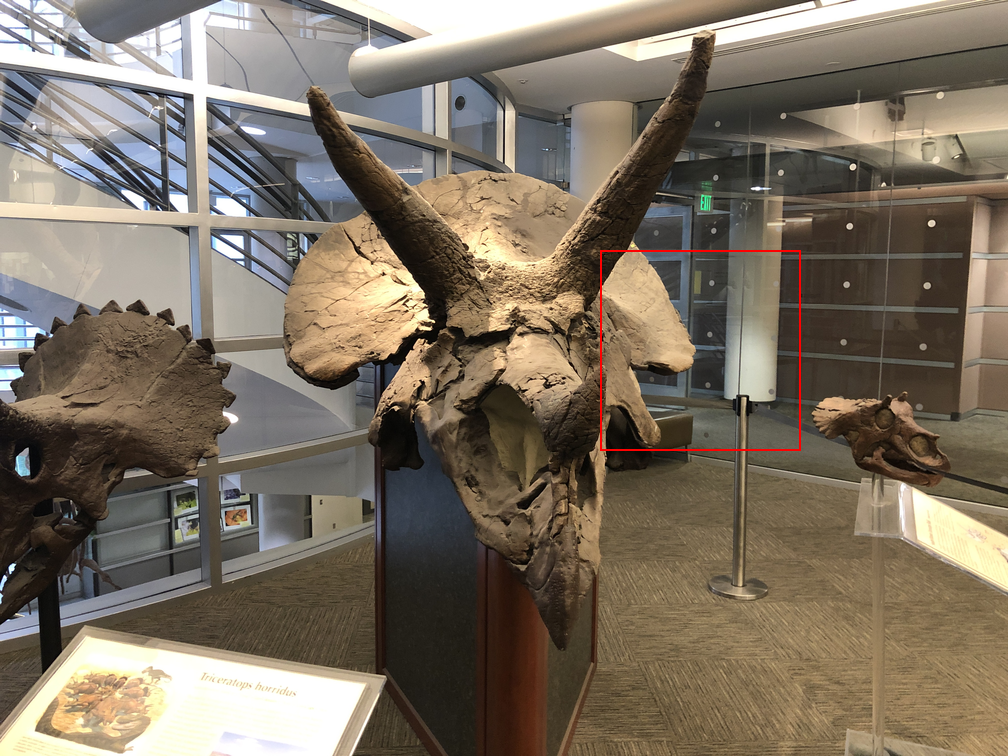} &
\includegraphics[width=0.2\linewidth]{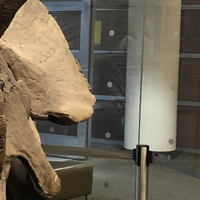} &
\includegraphics[width=0.2\linewidth]{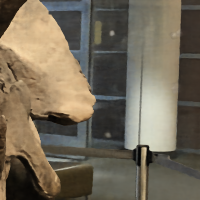} &
\includegraphics[width=0.2\linewidth]{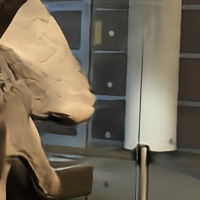} \\
\texttt{Horn} & (a) GT & (b) NeRF & (c) Ours \\
\includegraphics[width=0.2\linewidth]{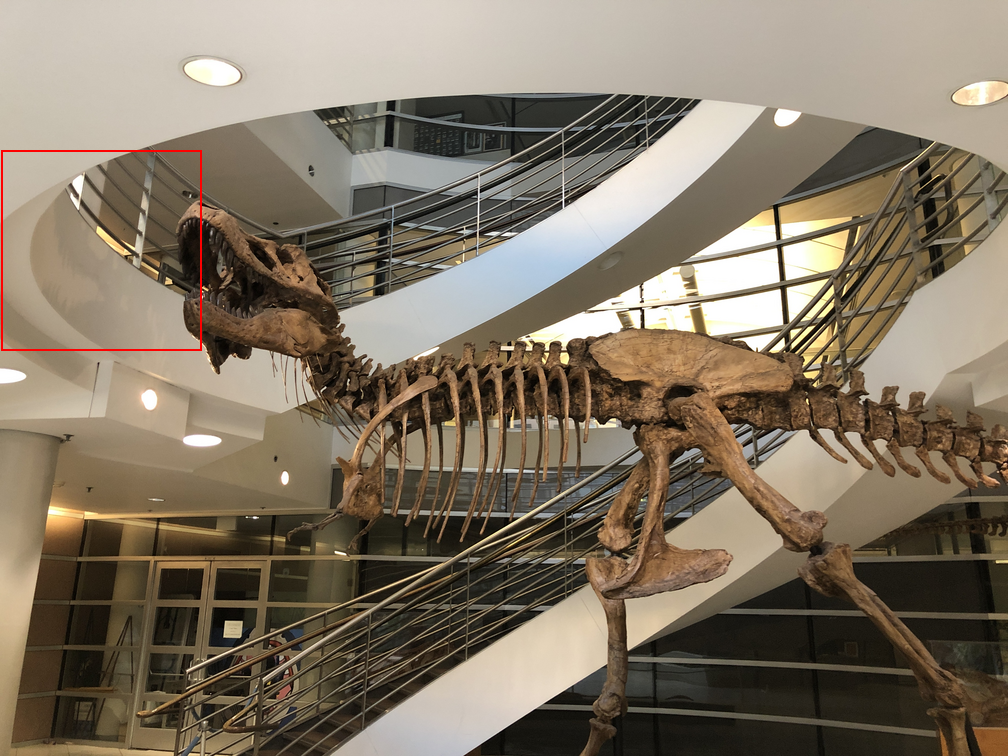} &
\includegraphics[width=0.2\linewidth]{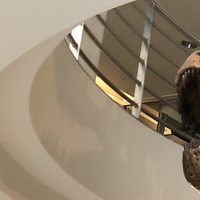} &
\includegraphics[width=0.2\linewidth]{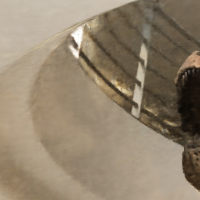} &
\includegraphics[width=0.2\linewidth]{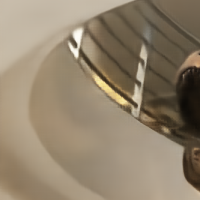} \\
\texttt{T-Rex} & (a) GT & (b) NeRF & (c) Ours \\
\end{tabular}}
\caption{More visual comparisons between our method and NeRF~\cite{nerf} (trained via NeRF-Pytorch~\cite{lin2020nerfpytorch}) on the synthetic $360^{\circ}$ (size: $800\times800\times3$) and real-world forward-facing scenes (size: $1008\times756\times3$). \textit{Best viewed in color and zoomed in}.}
\label{fig:more_visual_comparison}
\end{figure}